\documentclass{article}
\usepackage{iclr2027_conference,times}

\usepackage{amsmath,amsfonts,bm}

\def\eqref#1{equation~\ref{#1}}

\def\1{\bm{1}}

\DeclareMathAlphabet{\mathsfit}{\encodingdefault}{\sfdefault}{m}{sl}
\SetMathAlphabet{\mathsfit}{bold}{\encodingdefault}{\sfdefault}{bx}{n}

\usepackage{hyperref}
\hypersetup{hidelinks}
\usepackage{url}
\usepackage{graphicx}
\usepackage{booktabs}   
\usepackage{amsmath}    
\usepackage{placeins}
\usepackage{rotating}
\usepackage{multirow}
\usepackage{enumitem}
\usepackage{float}

\title{CHARTER: Auditing Reference Substitution in Hierarchical Compact-Evidence Evaluation for Computational Pathology}

\author{%
\makebox[5.3in][c]{%
\begin{tabular}{@{}c@{}}
\bfseries
Hyun Do Jung\textsuperscript{1}
\quad
Jungwon Choi\textsuperscript{2}
\quad
Soojung Choi\textsuperscript{1}
\quad
Yujin Oh\textsuperscript{1,\textdagger}
\quad
Hwiyoung Kim\textsuperscript{3,\textdagger}
\\[0.35em]
\normalfont\small
\textsuperscript{1}Yonsei University
\qquad
\textsuperscript{2}KAIST
\qquad
\textsuperscript{3}Hallym University Chuncheon Sacred Heart Hospital
\end{tabular}%
}%
}

\iclrfinalcopy

\begin{document}

\maketitle

\lhead{}
\renewcommand{\headrulewidth}{0pt}

\begingroup
\renewcommand{\thefootnote}{\textdagger}
\footnotetext{Corresponding authors: \texttt{yujinoh@yuhs.ac}, \texttt{hykim@hallym.or.kr}}
\endgroup

\begin{abstract}
In digital pathology, compact evidence is often used to explain or audit predictions made by whole-slide image multiple instance learning models. In hierarchical compact-evidence pipelines, candidate filtering introduces a strategy-specific candidate-conditioned prediction alongside the original full-bag prediction. If the evaluation reference changes while the intended target remains the original full-bag prediction, however, not only can the measured fidelity of the same compact evidence change, but comparisons between competing candidate strategies can also change. To make this dependence explicit, we introduce CHARTER, a reference-aware evaluation charter that asks researchers to \textbf{DECLARE} the intended target and reference, \textbf{QUANTIFY} candidate-induced prediction shift, and \textbf{AUDIT} the stability of comparative conclusions. Across the 15 comparisons in our main five-seed Random-$K$ audit, 4 showed determinate reversals; in a matched native-ranking stress test, the ACMIL comparison changed from REVERSED to PRESERVED. CHARTER turns otherwise implicit candidate-filtering and reference choices into an auditable evaluation specification, helping distinguish genuine preservation of the intended prediction from apparent gains induced by changing the prediction being explained.
\end{abstract}

\section{Introduction}
\label{sec:Introduction}

Whole-slide images (WSIs) are commonly represented as large bags of image patches and analyzed using multiple instance learning (MIL) to produce slide-level predictions~\citep{lu2021data,shao2021transmil}. Recent studies have attempted to provide compact evidence showing which patches support such predictions, or to use such evidence to interpret or audit model decisions beyond reporting predictive performance alone~\citep{kapse2024simil,hense2024xmil}. In particular, filtering informative patches into a compact subset can make model evidence more tractable to inspect and evaluate than examining the whole bag. Thus, the question of which subset best explains or preserves the model prediction itself becomes an important evaluation question.

Prior compact-evidence studies have addressed two related evaluation questions. One asks how well selected evidence preserves label-based downstream predictive performance, while another asks how well a compact subset preserves a fixed reference prediction in terms of fidelity or sufficiency~\citep{kapse2024simil,carter2019sis,deyoung2020eraser}. However, rather than selecting the final subset directly from the full bag, a hierarchical compact-evidence pipeline may first construct an intermediate candidate pool, within which further ranking or selection is performed to obtain the final subset. In such a pipeline, the original full-bag prediction is shared across candidate strategies, whereas the candidate-conditioned prediction is generated from each strategy's own candidate pool. Candidate-conditioned evaluation can therefore compare competing strategies against different, strategy-specific prediction references even when the intended target is the same original full-bag prediction.

Consider applying two candidate-filtering strategies to a single WSI. The candidate pool produced by the first strategy almost preserves the original full-bag prediction, whereas the candidate pool produced by the second strategy may shift the prediction substantially. After selecting a compact subset from each candidate pool and evaluating each subset against its corresponding candidate-pool prediction, the fidelities of the two strategies are calculated with respect to different prediction targets. In contrast, using the original full-bag prediction as a common reference allows the two strategies to be compared in terms of how well they preserve the same target. This does not mean that the candidate-conditioned prediction is an invalid reference; rather, using a strategy-specific reference may change the comparison when the intended target is the original full-bag prediction.

A concrete motivating instance is FOCI~\citep{jung2026compactrationalesfreemeasuring}, a hierarchical compact-evidence pipeline that applies candidate pre-filtering before final subset selection. Such an intermediate stage introduces a natural distinction between preserving the prediction from the candidate pool and preserving the original full-bag prediction. We therefore investigate this distinction as a measurement problem in hierarchical compact-evidence evaluation, holding the final subset prediction fixed while changing only its evaluation reference.

We audit reference substitution in a controlled setting using frozen WSI-MIL predictors by separately measuring candidate-induced prediction shift, reference-specific compact-subset fidelity, and the comparative gap between candidate strategies. The results show that the magnitude of the reference effect is not uniform across settings. In LUAD-TP53, large candidate-induced prediction shifts are accompanied by strong reference effects, such that the categorical conclusion of candidate comparisons can change, whereas the three-architecture CRC-MSI subset of the main audit shows comparatively small reference effects together with small candidate-induced shifts. Clean NSCLC shows heterogeneous regimes across architectures. We further observe determinate reversals under a controlled Random-$K$ evaluation, while the categorical outcome changes under a matched native-ranking stress test in a directly comparable setting, indicating that comparative conclusions can depend on the ranking protocol. Importantly, native ranking does not eliminate reference dependence: even when the categorical outcome is no longer reversed in the tested setting, a substantial reference-dependent comparative gap can remain.

Based on these results, rather than proposing a new selector or trainable method, we introduce CHARTER, a reference-aware evaluation charter for hierarchical compact-evidence evaluation. It comprises three actions: \textsc{Declare} the intended scientific target and prediction reference, \textsc{Quantify} candidate-induced prediction shift, and \textsc{Audit} the stability of comparative conclusions across reference choices and relevant ranking protocols. When the original full-bag prediction is the intended target, fidelity to that prediction should be reported as the primary result, while candidate-conditioned fidelity should be reported separately as a conditional result.

In summary, we make three contributions:
\begin{itemize}[leftmargin=1.5em, labelsep=0.5em, topsep=0.25em, itemsep=0.25em, parsep=0pt,
  partopsep=0pt]

\item \textbf{We formulate reference substitution as an explicit evaluation problem in hierarchical compact-evidence pipelines.}
We separate $X\rightarrow C\rightarrow S$ and distinguish the common original reference $q_0=f(X)$ from the strategy-specific candidate-conditioned reference $q_C=f(C)$, making explicit that both fidelity and candidate comparisons can change even for a fixed subset prediction $p=f(S)$.

\item \textbf{We show that reference substitution can change comparative conclusions.}
Under controlled Random-$K$, \textbf{4 of 15 comparisons were classified as REVERSED}; in a matched native-ranking stress test, ACMIL changed from \textbf{REVERSED to PRESERVED}, while a substantial reference-dependent comparative effect remained.

\item \textbf{We introduce CHARTER as an actionable reference-aware evaluation charter.}
CHARTER requires researchers to \textbf{DECLARE} the intended target and reference, \textbf{QUANTIFY} candidate-induced prediction shift, and \textbf{AUDIT} conclusion stability, making otherwise implicit reference choices auditable and separating original-target preservation from candidate-conditioned fidelity.

\end{itemize}

\section{Related Work}
\label{sec:RelatedWork}

\subsection{WSI-MIL and Compact Evidence}
WSI-MIL interpretability has expanded beyond attention visualization to include region-wise contributions, informative patch selection, compact evidence, and faithfulness evaluation. Additive MIL directly decomposes predictions into region-wise contributions~\citep{javed2022additivemil}, SI-MIL constructs an interpretable branch by selecting salient Top-$K$ patches~\citep{kapse2024simil}, and xMIL quantitatively evaluates the faithfulness of histopathology MIL explanations~\citep{hense2024xmil}. ReaMIL proposes an approach that directly learns compact evidence through a sufficiency-oriented objective~\citep{jung2026reamil}. Collectively, these methods construct, represent, or assess compact evidence once a prediction of interest has been specified. Our work instead asks which prediction should serve as the reference when the selected evidence is obtained through an intermediate candidate stage.

\subsection{Faithfulness and Subset-based Evaluation}
Sufficiency and faithfulness with respect to reduced or rationale-only inputs are well established in the compact-rationale literature. SIS identifies small input subsets that preserve the original model decision~\citep{carter2019sis}, while ERASER evaluates rationale faithfulness through sufficiency on rationale-only inputs and comprehensiveness under rationale removal~\citep{deyoung2020eraser}. Carton et al.\ further characterize these behaviors using fidelity curves as rationale content is progressively removed~\citep{carton2020rationales}. Related work in graph explanation considers compact subgraphs and distinguishes evaluation goals and fidelity notions~\citep{akkas2024gnnshap,amara2022graphframex}; more broadly, Jacovi and Goldberg emphasize that the faithfulness criterion itself must be stated explicitly~\citep{jacovi2020faithfulness}.

Together, these studies motivate evaluating reduced evidence against a specified model behavior. They do not, however, resolve the hierarchical case in which an intermediate candidate stage produces another prediction before the final compact subset is evaluated. Our focus is this narrower situation, where both the original full-bag prediction and the candidate-conditioned prediction can plausibly serve as evaluation references.

\subsection{Evaluation Sensitivity and Reference Substitution}
Measured faithfulness and method rankings can depend on the explanation metric or evaluation protocol~\citep{chan2022comparative}. Tomsett et al.\ show that differences in saliency-metric computation can alter measured fidelity and comparative rankings~\citep{tomsett2020saliency}, while MetaQuantus studies disagreement among explanation-quality estimators and the resulting ranking instability~\citep{hedstrom2023metaquantus}. Perturbation-based approaches, including xMIL, likewise assess changes in model output when evidence is removed or restricted~\citep{hense2024xmil}.

Our setting isolates a different source of variation. The fidelity criterion need not change; instead, each strategy's candidate pool induces its own candidate-conditioned prediction reference. We therefore hold each final compact-subset prediction fixed, evaluate it against both the common original full-bag prediction and the corresponding strategy-specific candidate-conditioned prediction, and examine how the resulting candidate-specific reference effects propagate to pairwise comparisons. In the literature we examined, this particular separation of a fixed subset prediction, a common original reference, and strategy-specific candidate-conditioned references has not been explicitly characterized in comparative evaluation.
\FloatBarrier

\FloatBarrier

\section{Reference Substitution in Hierarchical Compact-Evidence Evaluation}
\label{sec:reference_substitution}

\subsection{Hierarchical Compact-Evidence Pipeline}
\label{subsec:hierarchical_pipeline}

\begin{figure*}[!t]
    \centering
    \includegraphics[width=\textwidth]{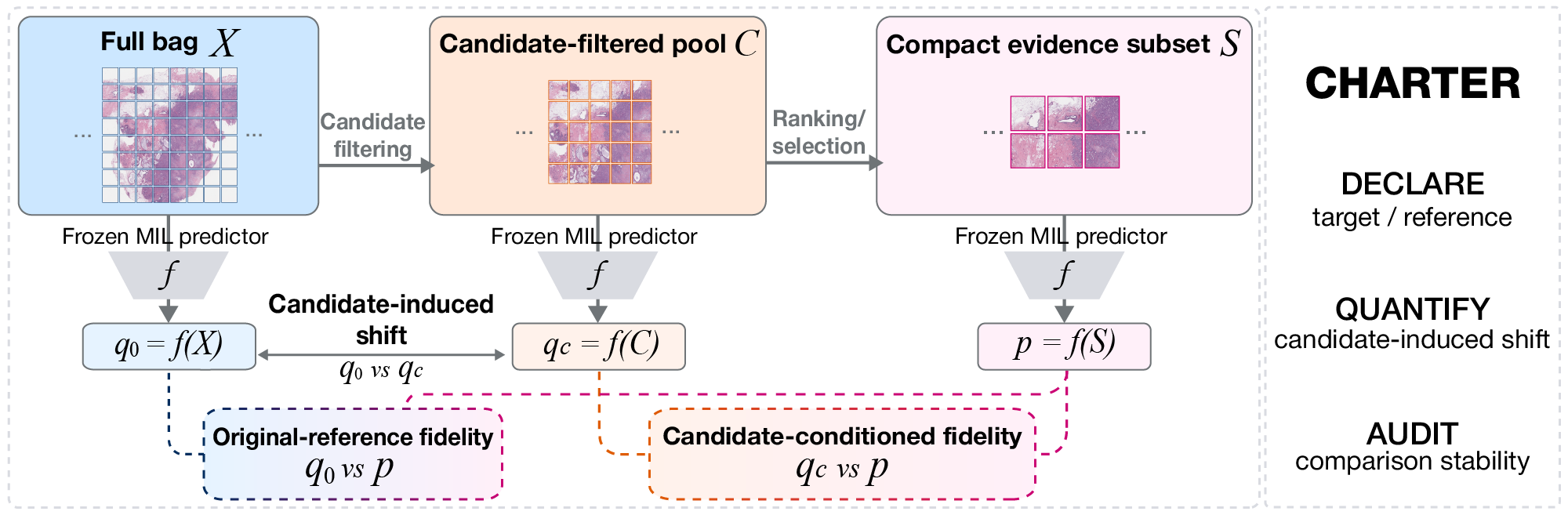}
    \caption{\textbf{Reference substitution in hierarchical compact-evidence evaluation and CHARTER.}
    The same frozen MIL predictor is applied to the full bag $X$, candidate pool $C$, and compact subset $S$, producing $q_0$, $q_C$, and $p$. Candidate filtering can move the prediction reference before compact-subset fidelity is assessed. CHARTER makes the intended target, reference shift, and comparative stability explicit.}
    \label{fig:overview}
\end{figure*}

In WSI-MIL, a slide is represented as a bag of patch-level representations \citep{ilse2018attention,lu2021data,shao2021transmil}, from which compact evidence can be selected or assessed for interpretation and auditing \citep{kapse2024simil,hense2024xmil}. We consider a hierarchical setting in which the full bag $X$ is first reduced to an intermediate candidate pool $C$, from which ranking or selection produces a final compact subset $S$. Critically, candidate filtering does not merely restrict the search space for $S$; it can also change the model output before the final subset is evaluated (Fig.~\ref{fig:overview}).

The same frozen MIL predictor $f$ is applied at all three stages, yielding the original full-bag prediction $q_0=f(X)$, candidate-conditioned prediction $q_C=f(C)$, and compact-subset prediction $p=f(S)$. Thus, $q_0$ and $q_C$ differ because the input bag changes, not because different predictors are used.

If the intended scientific target is the original full-bag prediction, the relevant fidelity question is how well $p$ preserves $q_0$. Conversely, $q_C$ is a legitimate reference when the candidate-conditioned prediction is itself the intended target. The problem arises when the intended target remains $q_0$, but the evaluation instead uses $q_C$ as the fidelity reference. We refer to this change in prediction reference as \emph{reference substitution}.

\subsection{Reference-Dependent Fidelity}
\label{subsec:reference_fidelity}

We quantify the prediction shift induced by candidate filtering as
\begin{equation}
\delta_C = D(q_0,q_C).
\end{equation}
For a fixed compact subset $S$ and its prediction $p=f(S)$, we define the two reference-specific fidelity errors as
\begin{equation}
e_{\mathrm{orig}} = D(p,q_0),
\qquad
e_{\mathrm{cond}} = D(p,q_C).
\end{equation}
These two errors evaluate the same compact-subset prediction against different references and therefore answer different fidelity questions; smaller values indicate higher fidelity to the corresponding reference. We use total variation (TV) distance for $D$ throughout our experiments. Importantly, when comparing original-reference and candidate-conditioned fidelity, the subset $S$ and prediction $p$ are held fixed; only the prediction reference changes.

\subsection{Comparative Conclusions under Reference Substitution}
\label{subsec:comparative_conclusions}
The same distinction extends to comparisons between candidate strategies: different candidate constructions can induce different reference effects, so changing the reference may also alter their relative comparison. For a candidate construction strategy $g$, let $C_g=g(X)$ denote its candidate pool, $S_g\subset C_g$ the compact subset obtained under a fixed ranking or selection protocol, $q_{C_g}=f(C_g)$ its candidate-conditioned prediction, and $p_g=f(S_g)$ its subset prediction. Fidelity success with respect to a reference $q$ requires $p_g$ to have the same predicted class as $q$ and satisfy $D(p_g,q)\leq\epsilon$. We denote the corresponding success rates by $\mathrm{Succ}^{\mathrm{orig}}_g$ and $\mathrm{Succ}^{\mathrm{cond}}_g$ and define the candidate-specific reference effect as
\begin{equation}
B_g =
\mathrm{Succ}^{\mathrm{cond}}_g -
\mathrm{Succ}^{\mathrm{orig}}_g .
\end{equation}
Thus, $B_g$ measures how much the aggregate fidelity-success rate of strategy $g$ changes when the reference is changed from $q_0$ to $q_{C_g}$.

For two candidate strategies $A$ and $B$, their reference-specific comparative gaps are
\begin{equation}
\Delta_{\mathrm{orig}}
=
\mathrm{Succ}^{\mathrm{orig}}_A -
\mathrm{Succ}^{\mathrm{orig}}_B,
\qquad
\Delta_{\mathrm{cond}}
=
\mathrm{Succ}^{\mathrm{cond}}_A -
\mathrm{Succ}^{\mathrm{cond}}_B .
\end{equation}
A positive gap favors strategy $A$, whereas a negative gap favors strategy $B$. We quantify the change in this comparison directly as
\begin{equation}
R_{A,B}
=
\Delta_{\mathrm{cond}}
-
\Delta_{\mathrm{orig}}
=
B_A-B_B .
\end{equation}

A change in comparative gap does not by itself imply a reversal in candidate ordering. We therefore classify a comparison as \textsc{Preserved} when the ordering is determinate under both references and retains the same direction, \textsc{Reversed} when it is determinate under both references but changes direction, and \textsc{Ranking-Indeterminate} when the ordering cannot be reliably determined under at least one reference. Consequently, a raw sign change alone is not sufficient to declare a ranking reversal.

A metric-based diagnostic and a sufficient condition for agreement of the discrete fidelity verdict are derived in Appendix~\ref{app:diagnostic}.

\section{Experimental Design}
\label{sec:experiments}

\subsection{Evaluation Scope}
\label{subsec:evaluation_scope}

To assess whether reference-substitution effects depend on task, MIL architecture, or pathology encoder, we audit these three axes separately. The main Random-$K$ audit contains every task--architecture pair with complete five-seed evaluations when the analysis was defined: nine MIL architectures on clean NSCLC and ABMIL~\citep{ilse2018attention}, TransMIL~\citep{shao2021transmil}, and ACMIL~\citep{zhang2024attention} on LUAD-TP53 and CRC-MSI (15 pairs), all derived from TCGA cohorts~\citep{weinstein2013cancer}. Selection was based on evaluation availability rather than the observed outcome.

We broaden this audit under a common $C_{\mathrm{L2}}$ protocol to a 12-task $\times$ 9-architecture grid (108 cells), with 70 validated five-seed estimates. Figure~3A shows the complete common-coverage subset of 10 tasks $\times$ 3 MIL architectures (30 cells); PANDA~\citep{bulten2022artificial} and CAMELYON+~\citep{ling2025camelyon} have no validated estimate and are excluded from the reported breadth claims.

For the encoder axis, the planned grid covers three tasks, five MIL architectures, and four pathology encoders (60 cells). Of 24 validated estimates, eight task--MIL settings are matched between UNI2-h~\citep{chen2024towards} and Phikon-v2~\citep{filiot2024phikon_v2}, yielding the 16 estimates shown in Figure~3B. Each predictor uses five fixed seeds, $\{0,1,42,1337,2025\}$, and remains frozen during evaluation; cohort, split, analysis-unit, and architecture details are provided in Appendices~B--C.

\subsection{Candidate Construction and Ranking}
\label{subsec:candidate_ranking}

Candidate construction and within-candidate ranking are separate evaluation axes: the former maps the full bag $X$ to a candidate pool $C$, whereas the latter produces the compact subset $S$. The primary candidate constructions are $C_{\mathrm{L2}}$, which selects the top-$M$ candidates by L2 norm in the frozen predictor's token representation space, and $C_{\mathrm{RAND}}$, an identity-keyed deterministic uniform random construction. For bags with fewer than $M$ tiles, the effective candidate budget is $M_c=\min(M,N)$; both candidate constructions then retain the full bag, and subset selection uses $K_c=\min(K,M_c)$.

Random-$K$ is the primary controlled ranker, while the tested predictor-derived native-ranking signals serve as a matched stress test of conclusion stability under model-derived ranking. For each slide, native scores are computed once from the full bag using the same frozen predictor and then restricted to the corresponding candidate pool before top-$K$ selection. ABMIL and CLAM~\citep{ilse2018attention,lu2021data} use pre-softmax gated-attention logits, ACMIL uses the mean of its five branch logits, and TransMIL uses the dot product between the pre-encoder tile embedding and learned class-token vector. Candidate-construction and ranking implementation details are provided in Appendix~\ref{app:candidate_specs}.

\subsection{Evaluation and Statistical Protocol}
\label{subsec:evaluation_statistics}

Unless otherwise specified, the default operating point is $M=1024$, $K=32$, and $\epsilon=0.05$. The same frozen predictor produces $q_0$, $q_C$, and $p$; fidelity success requires predicted-class agreement with the reference and $\mathrm{TV}(p,q)\leq\epsilon$, with $p$ held fixed when the reference is changed.

Inference uses patient-level resampling where patient identifiers are available; for PANDA, which lacks patient identifiers, the planned evaluation uses slide-level resampling. Predictor seeds are treated as repeated model realizations rather than independent observations. Pairwise reference effects $R_{A,B}$ are computed directly within each bootstrap resample. A reference-specific ordering is determinate only when its 95\% confidence interval excludes zero and its absolute gap exceeds the fixed tolerance $\tau=0.05$; otherwise it is \textsc{Ranking-Indeterminate}. Determinate orderings retaining or changing direction are labeled \textsc{Preserved} or \textsc{Reversed}, respectively; raw sign changes alone do not constitute reversals. Further statistical and aggregation details are provided in Appendix~\ref{app:stats}.

Provenance and validation controls are described in Appendix~\ref{app:provenance}.

\section{Results}
\label{sec:results}

\subsection{When Does Reference Substitution Materially Change Fidelity?}
\label{subsec:results_regimes}

We first ask when reference substitution materially changes measured fidelity. We examine the candidate-induced prediction shift $\delta_C$ together with the corresponding candidate-specific reference effect $B_g$. The effect was not uniform across tasks, architectures, or candidate constructions: LUAD-TP53 showed a strong-effect regime, the three-architecture CRC-MSI subset of the main audit a comparatively small-effect regime, and clean NSCLC substantial architecture-dependent variation.

Table~\ref{tab:regime_summary} provides a compact summary of these regimes at the default operating point, including candidate-induced shifts, candidate-specific reference effects, and the corresponding pairwise outcomes.

LUAD-TP53 showed the strongest reference-substitution regime: under $C_{\mathrm{L2}}$, both candidate-induced shifts and reference effects were substantially larger than under $C_{\mathrm{RAND}}$, with all three $B_g$ estimates significant (Table~\ref{tab:regime_summary}). Within the three-architecture CRC-MSI subset of the main audit, reference effects were comparatively small, although it was not completely null: ABMIL under $C_{\mathrm{L2}}$ and ACMIL under $C_{\mathrm{RAND}}$ retained significant reference effects. Clean NSCLC exhibited substantial architecture-dependent variation; $C_{\mathrm{L2}}$ produced significant $B_g$ estimates across all nine architectures, whereas under $C_{\mathrm{RAND}}$ only MaxMIL was significant.

Across the three tasks, candidate-induced prediction shift and candidate-specific reference effect were therefore related but distinct quantities. Settings with larger $\delta_C$ tended to show larger $B_g$, but whether these fidelity changes also altered candidate ordering requires a separate pairwise comparison.

\begin{table}[!b]
\centering
\caption{\textbf{Reference-substitution regimes under Random-$K$.} Candidate-specific effects and pairwise outcomes under the fixed tolerance rule for $C_{\mathrm{L2}}$ versus $C_{\mathrm{RAND}}$.}
\label{tab:regime_summary}
\begingroup
\small
\setlength{\tabcolsep}{3.2pt}
\renewcommand{\arraystretch}{1.08}
\begin{tabular*}{\linewidth}{@{\extracolsep{\fill}}lclcccc@{}}
\toprule
 & & \multicolumn{4}{c}{Candidate-specific quantities} & Pairwise outcome \\
\cmidrule(lr){3-6}\cmidrule(l){7-7}
Task & MILs & Candidate & \shortstack{Prediction shift\\$\delta_C$ (min--max)} & \shortstack{Fidelity effect\\$B_g$ (min--max)} & \shortstack{Sig.\ $B_g$\\MILs} & R / P / I \\
\midrule
\multirow{2}{*}{LUAD-TP53} & \multirow{2}{*}{3} & $C_{\mathrm{L2}}$ & $0.1585$--$0.2276$ & $0.1612$--$0.4537$ & 3/3 & \multirow{2}{*}{2 / 0 / 1} \\
 & & $C_{\mathrm{RAND}}$ & $0.0207$--$0.0287$ & $0.006$--$0.0179$ & 0/3 & \\
\addlinespace[3pt]
\multirow{2}{*}{CRC-MSI} & \multirow{2}{*}{3} & $C_{\mathrm{L2}}$ & $0.0124$--$0.018$ & $0.0123$--$0.037$ & 1/3 & \multirow{2}{*}{0 / 1 / 2} \\
 & & $C_{\mathrm{RAND}}$ & $0.0066$--$0.0157$ & $0.0000$--$0.0173$ & 1/3 & \\
\addlinespace[3pt]
\multirow{2}{*}{Clean NSCLC} & \multirow{2}{*}{9} & $C_{\mathrm{L2}}$ & $0.0179$--$0.0636$ & $0.0232$--$0.2428$ & 9/9 & \multirow{2}{*}{2 / 1 / 6} \\
 & & $C_{\mathrm{RAND}}$ & $0.0033$--$0.0141$ & $0.0000$--$0.0358$ & 1/9 & \\
\bottomrule
\end{tabular*}
\par\vspace{3pt}
\begin{minipage}{\linewidth}
\footnotesize
Ranges are across the listed MILs after five-seed pooling. ``Sig.'' indicates a pointwise 95\% bootstrap confidence interval for $B_g$ excluding zero. R/P/I = reversed / preserved / ranking-indeterminate; significance of $B_g$ alone does not imply reversal.
\end{minipage}
\endgroup
\end{table}

\begin{figure*}[!t]
    \centering
    \includegraphics[width=\textwidth]{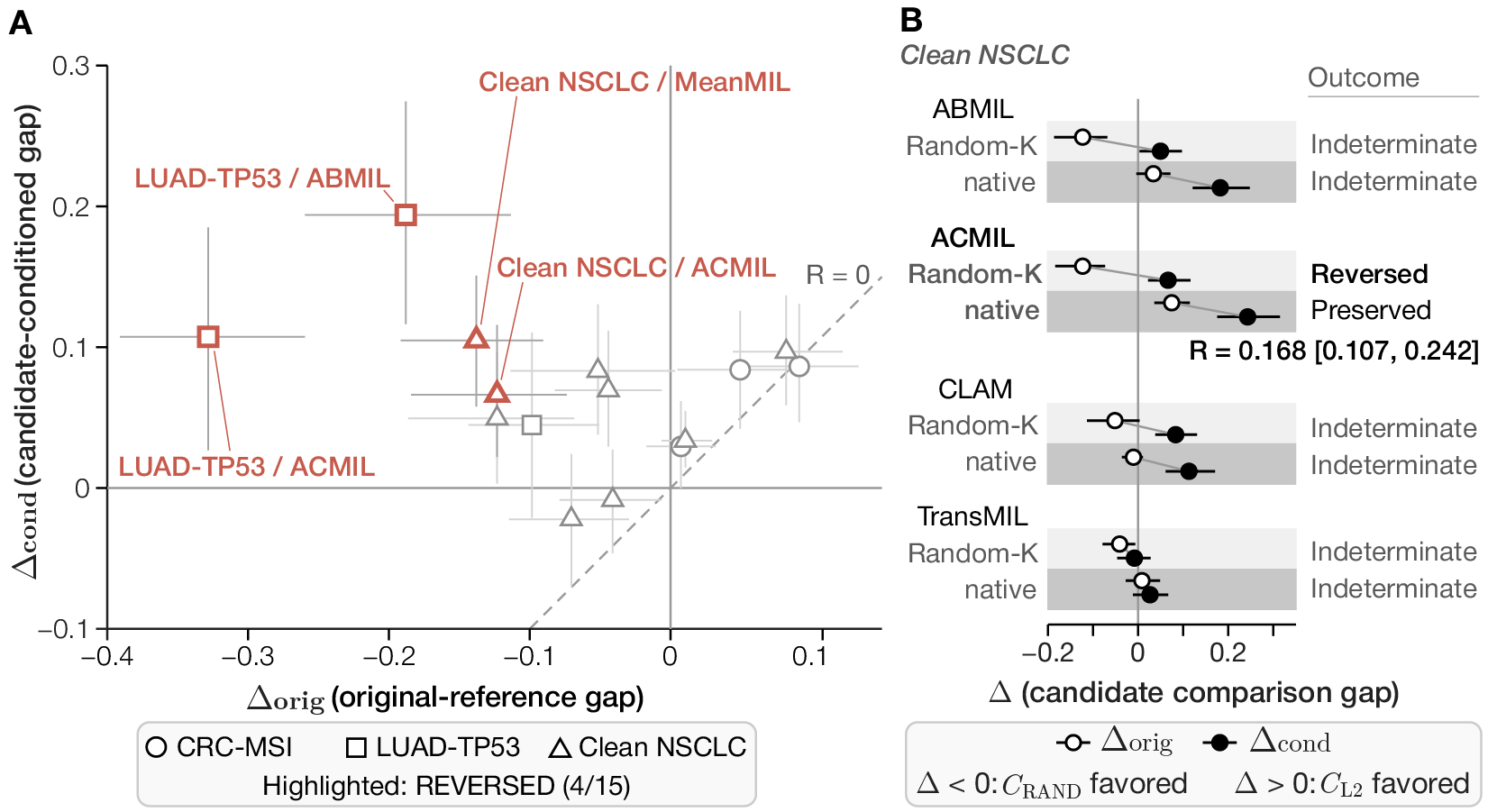}
    \caption{
    \textbf{Reference substitution can change comparative conclusions, with outcomes depending on the ranking protocol.}
    \textbf{A}, candidate gaps under the original and candidate-conditioned references; error bars denote 95\% confidence intervals. \textsc{Reversed}/\textsc{Preserved} outcomes require a determinate gap under both references (95\% CI excluding zero and $|\Delta|>\tau=0.05$); otherwise the comparison is \textsc{Ranking-Indeterminate}.
    \textbf{B}, matched comparison of controlled Random-$K$ and the tested predictor-derived native-ranking signals.
    }
    \label{fig:comparative_effect}
\end{figure*}

\subsection{Can Reference Substitution Change Comparative Conclusions?}
\label{subsec:results_comparative}

We next ask whether these reference-specific fidelity changes can alter the pairwise comparison between $C_{\mathrm{L2}}$ and $C_{\mathrm{RAND}}$. Among the 15 comparisons in the main five-seed Random-$K$ audit, four were REVERSED, two were PRESERVED, and nine were RANKING-INDETERMINATE. All four reversals occurred in the same direction: $C_{\mathrm{RAND}}$ had the larger fidelity-success rate under the shared full-bag reference, whereas $C_{\mathrm{L2}}$ had the larger rate under the respective candidate-conditioned references. In particular, for LUAD-TP53 with ABMIL, the comparison changed from $\Delta_{\mathrm{orig}}=-0.188$ to $\Delta_{\mathrm{cond}}=0.194$, with $R=0.382\,[0.296,0.463]$, resulting in a REVERSED outcome. Clean NSCLC with MeanMIL similarly changed from $-0.138$ to $0.105$, with $R=0.243\,[0.175,0.310]$, and was also REVERSED. In contrast, a significant direct $R$ does not necessarily imply a reversal: LUAD-TP53 with TransMIL had $R=0.143\,[0.081,0.209]$, but its outcome remained RANKING-INDETERMINATE. The remaining comparisons are shown in Fig.~\ref{fig:comparative_effect}A.

Random-$K$ removes architecture-specific subset-ranking signals; the matched native-ranking analysis tests whether reference dependence persists when predictor-derived ranking is restored. Under the matched predictor-derived rankings, none of the four clean-NSCLC comparisons was classified as \textsc{Reversed}. ACMIL changed from \textsc{Reversed} under Random-$K$ to \textsc{Preserved}, whereas ABMIL, CLAM, and TransMIL remained \textsc{Ranking-Indeterminate}. Reference dependence nevertheless remained measurable: the direct effect $R$ had a positive pointwise 95\% confidence interval excluding zero for ABMIL, ACMIL, and CLAM, with ACMIL showing $R=0.168$ [0.107, 0.242]. Thus, the absence of a determinate reversal under predictor-derived ranking does not imply that the candidate comparison is insensitive to the prediction reference. The matched ranking-protocol results are shown in Fig.~\ref{fig:comparative_effect}B, with exact numerical comparisons in Appendix Table~\ref{tab:app-p3p4-full}.

\subsection{How Broadly Does the Effect Appear Across Tasks and MIL Architectures?}

We next ask whether the reference-substitution effect observed in the main audit extends across a broader range of tasks and MIL architectures. The broader task--architecture grid contains 70 validated five-seed estimates among 108 planned cells under the common $C_{\mathrm{L2}}$ protocol. To make the cross-task comparison directly readable, Figure~3A shows the complete common-coverage subset of 10 tasks evaluated with ABMIL, TransMIL, and ACMIL (30 cells); this subset is defined by shared coverage rather than by the observed outcome.

The resulting pattern is heterogeneous rather than concentrated in a single task or MIL family. Substantial positive reference effects and determinate reversals coexist with small or ranking-indeterminate effects, and these differences can appear within the same task. For example, on LUAD-KRAS, ABMIL showed $R=0.291$ [0.191, 0.383] with a \textsc{Reversed} outcome, whereas TransMIL showed $R=0.000$ [0.000, 0.000] and remained \textsc{Ranking-Indeterminate}. Thus, the broader audit suggests that reference substitution is not tied to one particular task or architecture, while its magnitude and categorical consequence remain strongly setting-dependent.

\begin{figure*}[!htbp]
    \centering
    \includegraphics[width=\textwidth]{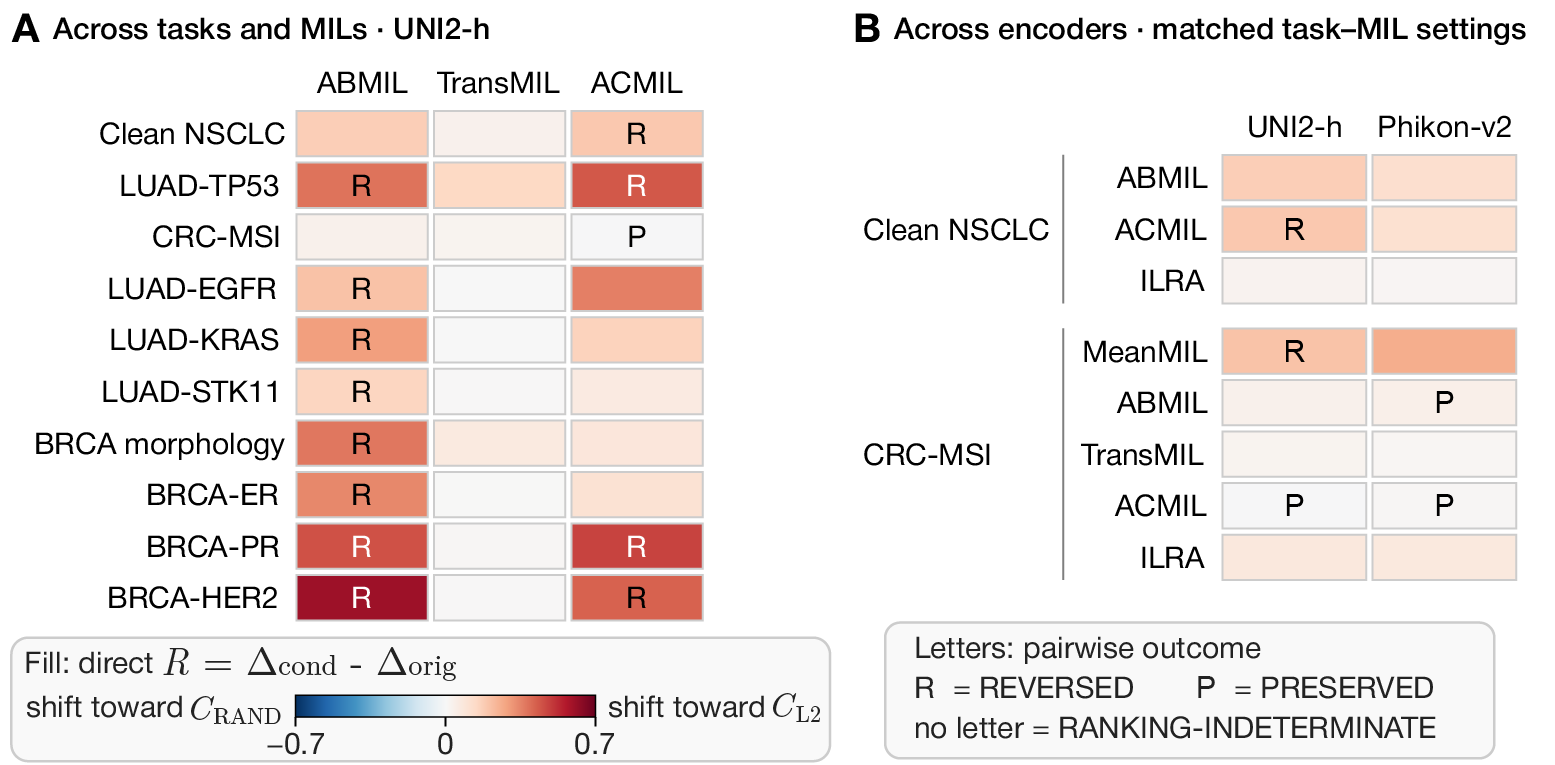}
    \caption{
    \textbf{Breadth of the reference-substitution effect across tasks, MIL architectures, and pathology encoders.}
    The broader audits contain 70 validated estimates among 108 planned task--MIL cells and 24 validated estimates among 60 planned encoder cells.
    \textbf{A}, the complete common-coverage subset of 10 tasks $\times$ 3 MIL architectures (ABMIL, TransMIL, and ACMIL; 30 cells).
    \textbf{B}, all eight task--MIL settings directly matched between UNI2-h and Phikon-v2 (16 estimates).
    Fill encodes the signed direct pairwise reference effect $R=\Delta_{\mathrm{cond}}-\Delta_{\mathrm{orig}}$ on a shared scale; R/P denote \textsc{Reversed}/\textsc{Preserved}, and cells without a letter are \textsc{Ranking-Indeterminate}.
    Displayed subsets are defined by common coverage or encoder matching rather than by the observed outcome.
    }
    \label{fig:breadth_matrix}
\end{figure*}

\subsection{How Does the Effect Vary Across Pathology Encoders?}

We finally ask whether reference-dependent effects remain observable when the pathology feature encoder is changed. Among the 24 validated estimates in the broader encoder grid, eight task--MIL settings are directly matched between UNI2-h and Phikon-v2, yielding the 16 estimates shown in Figure~3B.

The matched comparisons show that changing the encoder can alter both the magnitude and the categorical consequence of the reference effect without consistently removing the effect itself. For CRC-MSI with MeanMIL, for example, UNI2-h showed $R=0.203$ [0.133, 0.282] with a \textsc{Reversed} outcome, whereas Phikon-v2 showed $R=0.257$ [0.178, 0.341] while the comparison remained \textsc{Ranking-Indeterminate}. Across the eight matched Phikon-v2 settings, outcomes were 0 \textsc{Reversed}, 2 \textsc{Preserved}, and 6 \textsc{Ranking-Indeterminate}, showing that continuous reference dependence can persist while categorical consequences remain encoder-dependent. The complete main-panel match is limited to UNI2-h and Phikon-v2; a single CRC-MSI/ABMIL three-encoder anchor is reported in Appendix~\ref{app:additional_results} and does not establish broad encoder invariance.

\section{CHARTER: A Reference-Aware Evaluation Charter}

CHARTER (Candidate-aware Hierarchical Assessment of References, Targets, and Evaluation Reporting) summarizes the preceding audit into three evaluation and reporting actions.

\paragraph{DECLARE the Intended Target and Reference.}
The previous formulation shows that reference substitution is not a simple implementation choice, but may change the scientific question being evaluated. Thus, the intended prediction target and evaluation reference should be explicitly declared. If the full-bag prediction $q_0$ is the intended target, fidelity to $q_0$ should be reported as the primary result, while the use of $q_C$ should be explicitly stated when the candidate-conditioned prediction is the intended target. CHARTER does not force a single specific reference in all situations, but requires alignment between the intended target and the evaluation reference.

\paragraph{QUANTIFY Candidate-Induced Prediction Shift.}
Declaring the reference alone is not enough; how much the intermediate candidate filtering itself changes the prediction should also be reported. For each candidate strategy, the candidate-induced shift $\delta_C$, together with original-reference and candidate-conditioned fidelity, makes this change explicit. This allows us to distinguish regimes in which candidate filtering approximately preserves the original prediction from regimes in which reference substitution materially changes the measured fidelity.

\paragraph{AUDIT Conclusion Stability.}
Finally, candidate comparisons should examine $\Delta_{\mathrm{orig}}$, $\Delta_{\mathrm{cond}}$, the direct pairwise reference effect $R_{A,B}$ with its uncertainty, and the final ranking label together. A nonzero reference effect does not by itself imply a determinate reversal. When the scientific claim is intended to span ranking protocols, whether the comparative conclusion remains stable should also be audited across the relevant protocols. The purpose of this stage is not to designate a particular ranker as the correct choice, but to explicitly show how dependent the reported conclusion is on the evaluation specification.

\section{Discussion and Limitations}
\label{sec:Discussion}

The main observation of this study is that the fidelity of compact evidence is not determined only by the selected subset itself, but also by the prediction reference against which that subset is evaluated. In hierarchical compact-evidence pipelines, this matters because candidate construction can induce strategy-specific references, so competing strategies may be judged against different preservation targets even under the same fidelity criterion. Our results show that this choice can affect both measured fidelity and comparative conclusions. The matched native-ranking result further shows that ranking stability and reference stability are distinct: ACMIL was \textsc{Preserved} under predictor-derived ranking while retaining a substantial reference-dependent comparative gap ($R=0.168$ [0.107, 0.242]). Thus, the absence of a reversal does not by itself imply that the evaluation is insensitive to the chosen prediction reference. The candidate-conditioned prediction $q_C$ is not intrinsically an invalid reference; the issue is whether the intended scientific target and
evaluation reference are aligned.

This study establishes the existence and materiality of reference substitution in controlled WSI-MIL settings, not its field-wide prevalence. The native-ranking stress test shows protocol dependence without identifying a generally correct ranker. Prediction fidelity measures preservation of the frozen predictor, not biological or clinical validity; likewise, the sufficient-condition diagnostic is one-sided, so failure to certify does not imply disagreement or instability.

\section{Conclusion}
\label{sec:conclusion}

In hierarchical compact-evidence evaluation, intermediate candidate filtering can change the prediction reference before compact evidence is evaluated. We show that this can materially alter measured fidelity and, in some settings, comparative conclusions. CHARTER therefore calls for aligning the intended target and evaluation reference and auditing the consequences of that choice, without treating candidate-conditioned predictions as intrinsically invalid.

\section*{AI Use Statement}
Generative AI tools were used as assistive tools for research planning and methodological discussion, implementation and workflow support, discussion and cross-checking of experimental results and formal derivations, and manuscript and figure drafting and editing. All experiments and quantitative analyses were carried out through the authors' documented computational pipelines. The authors reviewed and verified all AI-assisted material and take responsibility for the scientific decisions, interpretations, claims, and final content of the manuscript.

\bibliography{iclr2027_conference}

@article{lu2021data,
  title     = {Data-efficient and weakly supervised computational pathology on whole-slide images},
  author    = {Lu, Ming Y and Williamson, Drew FK and Chen, Tiffany Y and Chen, Richard J and Barbieri, Matteo and Mahmood, Faisal},
  journal   = {Nature biomedical engineering},
  volume    = {5},
  number    = {6},
  pages     = {555--570},
  year      = {2021},
  publisher = {Nature Publishing Group UK London}
}

@inproceedings{shao2021transmil,
  title     = {{TransMIL}: Transformer based Correlated Multiple Instance Learning for Whole Slide Image Classification},
  author    = {Shao, Zhuchen and Bian, Hao and Chen, Yang and Wang, Yifeng and Zhang, Jian and Ji, Xiangyang and Zhang, Yongbing},
  booktitle = {Advances in Neural Information Processing Systems},
  editor    = {M. Ranzato and A. Beygelzimer and Y. Dauphin and P.S. Liang and J. Wortman Vaughan},
  volume    = {34},
  pages     = {2136--2147},
  year      = {2021},
  publisher = {Curran Associates, Inc.}
}

@inproceedings{kapse2024simil,
  title     = {{SI-MIL}: Taming Deep {MIL} for Self-Interpretability in Gigapixel Histopathology},
  author    = {Kapse, Saarthak and Pati, Pushpak and Das, Srijan and Zhang, Jingwei and Chen, Chao and Vakalopoulou, Maria and Saltz, Joel and Samaras, Dimitris and Gupta, Rajarsi R. and Prasanna, Prateek},
  booktitle = {Proceedings of the IEEE/CVF Conference on Computer Vision and Pattern Recognition (CVPR)},
  pages     = {11226--11237},
  year      = {2024},
}

@inproceedings{hense2024xmil,
  author = {Hense, Julius and Jamshidi Idaji, Mina and Eberle, Oliver and Schnake, Thomas and Dippel, Jonas and Ciernik, Laure and Buchstab, Oliver and Mock, Andreas and Klauschen, Frederick and M\"{u}ller, Klaus-Robert},
  booktitle = {Advances in Neural Information Processing Systems},
  editor = {A. Globerson and L. Mackey and D. Belgrave and A. Fan and U. Paquet and J. Tomczak and C. Zhang},
  pages = {8300--8328},
  publisher = {Curran Associates, Inc.},
  title = {{xMIL}: Insightful Explanations for Multiple Instance Learning in Histopathology},
  volume = {37},
  year = {2024}
}

@inproceedings{deyoung2020eraser,
  title     = {{ERASER}: {A} Benchmark to Evaluate Rationalized {NLP} Models},
  author    = {DeYoung, Jay and Jain, Sarthak and Rajani, Nazneen Fatema and Lehman, Eric and Xiong, Caiming and Socher, Richard and Wallace, Byron C.},
  booktitle = {Proceedings of the 58th Annual Meeting of the Association for Computational Linguistics},
  pages     = {4443--4458},
  year      = {2020},
  doi       = {10.18653/v1/2020.acl-main.408}
}

@misc{jung2026compactrationalesfreemeasuring,
  title         = {Are Compact Rationales Free? Measuring Tile Selection Headroom in Frozen {WSI-MIL}},
  author        = {Hyun Do Jung and Jungwon Choi and Soojung Choi and Yujin Oh and Hwiyoung Kim},
  year          = {2026},
  eprint        = {2605.12575},
  archivePrefix = {arXiv},
  primaryClass  = {eess.IV},
  url={https://arxiv.org/abs/2605.12575}
}

@inproceedings{javed2022additivemil,
  title={Additive {MIL}: Intrinsically Interpretable Multiple Instance Learning for Pathology},
  author={Syed Ashar Javed and Dinkar Juyal and Harshith Padigela and Amaro Taylor-Weiner and Limin Yu and aaditya prakash},
  booktitle={Advances in Neural Information Processing Systems},
  editor={Alice H. Oh and Alekh Agarwal and Danielle Belgrave and Kyunghyun Cho},
  year={2022},
}

@inproceedings{jung2026reamil,
    author    = {Jung, Hyun Do and Choi, Jungwon and Kim, Hwiyoung},
    title     = {{ReaMIL}: Reasoning- and Evidence-Aware Multiple Instance Learning for Whole-Slide Histopathology},
    booktitle = {Proceedings of the IEEE/CVF Winter Conference on Applications of Computer Vision (WACV) Workshops},
    month     = {March},
    year      = {2026},
    pages     = {40-45},
}

@inproceedings{carter2019sis,
  title     = {What Made You Do This? Understanding Black-box Decisions with Sufficient Input Subsets},
  author    = {Carter, Brandon and Mueller, Jonas and Jain, Siddhartha and Gifford, David},
  booktitle = {Proceedings of the Twenty-Second International Conference on Artificial Intelligence and Statistics},
  pages     = {567--576},
  year      = {2019},
  volume    = {89},
  series    = {Proceedings of Machine Learning Research},
  publisher = {PMLR}
}

@inproceedings{carton2020rationales,
  title     = {Evaluating and Characterizing Human Rationales},
  author    = {Carton, Samuel and Rathore, Anirudh and Tan, Chenhao},
  booktitle = {Proceedings of the 2020 Conference on Empirical Methods in Natural Language Processing (EMNLP)},
  pages     = {9294--9307},
  year      = {2020},
  doi       = {10.18653/v1/2020.emnlp-main.747}
}

@inproceedings{akkas2024gnnshap,
  title     = {{GNNShap}: Scalable and Accurate {GNN} Explanation Using Shapley Values},
  author    = {Akkas, Selahattin and Azad, Ariful},
  booktitle = {Proceedings of the ACM Web Conference 2024},
  pages     = {827--838},
  year      = {2024},
  doi       = {10.1145/3589334.3645599}
}

@inproceedings{amara2022graphframex,
  title     = {{GraphFramEx}: Towards Systematic Evaluation of Explainability Methods for Graph Neural Networks},
  author    = {Amara, Kenza and Ying, Zhitao and Zhang, Zitao and Han, Zhichao and Zhao, Yang and Shan, Yinan and Brandes, Ulrik and Schemm, Sebastian and Zhang, Ce},
  booktitle = {Proceedings of the First Learning on Graphs Conference},
  pages     = {44:1--44:23},
  volume    = {198},
  series    = {Proceedings of Machine Learning Research},
  publisher = {PMLR},
  year      = {2022}
}

@inproceedings{jacovi2020faithfulness,
  title     = {Towards Faithfully Interpretable {NLP} Systems: How Should We Define and Evaluate Faithfulness?},
  author    = {Jacovi, Alon and Goldberg, Yoav},
  booktitle = {Proceedings of the 58th Annual Meeting of the Association for Computational Linguistics},
  pages     = {4198--4205},
  month     = {jul},
  year      = {2020},
  doi       = {10.18653/v1/2020.acl-main.386}
}

@inproceedings{chan2022comparative,
    title = "A Comparative Study of Faithfulness Metrics for Model Interpretability Methods",
    author = "Chan, Chun Sik and Kong, Huanqi and Guanqing, Liang",
    booktitle = "Proceedings of the 60th Annual Meeting of the Association for Computational Linguistics (Volume 1: Long Papers)",
    month = may,
    year = "2022",
    address = "Dublin, Ireland",
    publisher = "Association for Computational Linguistics",
    doi = "10.18653/v1/2022.acl-long.345",
    pages = "5029--5038"
}

@inproceedings{tomsett2020saliency,
  title     = {Sanity Checks for Saliency Metrics},
  author    = {Tomsett, Richard and Harborne, Dan and Chakraborty, Supriyo and Gurram, Prudhvi and Preece, Alun D.},
  booktitle = {Proceedings of the AAAI Conference on Artificial Intelligence},
  volume    = {34},
  number    = {04},
  pages     = {6021--6029},
  year      = {2020},
  doi       = {10.1609/aaai.v34i04.6064}
}

@article{hedstrom2023metaquantus,
  title   = {The Meta-Evaluation Problem in Explainable {AI}: Identifying Reliable Estimators with MetaQuantus},
  author  = {Hedstr{\"o}m, Anna and Bommer, Philine and Wickstr{\o}m, Kristoffer K. and Samek, Wojciech and Lapuschkin, Sebastian and H{\"o}hne, Marina M.-C.},
  journal = {Transactions on Machine Learning Research},
  year    = {2023},
  issn    = {2835-8856}
}

@inproceedings{ilse2018attention,
  title        = {Attention-based deep multiple instance learning},
  author       = {Ilse, Maximilian and Tomczak, Jakub and Welling, Max},
  booktitle    = {Proceedings of the 35th International Conference on Machine Learning},
  pages        = {2127--2136},
  year         = {2018},
  organization = {PMLR}
}

@article{chen2024towards,
   author = {Chen, Richard J. and Ding, Tong and Lu, Ming Y. and Williamson, Drew F. K. and Jaume, Guillaume and Song, Andrew H. and Chen, Bowen and Zhang, Andrew and Shao, Daniel and Shaban, Muhammad and Williams, Mane and Oldenburg, Lukas and Weishaupt, Luca L. and Wang, Judy J. and Vaidya, Anurag and Le, Long Phi and Gerber, Georg and Sahai, Sharifa and Williams, Walt and Mahmood, Faisal},
   title = {Towards a general-purpose foundation model for computational pathology},
   journal = {Nature Medicine},
   volume = {30},
   number = {3},
   pages = {850--862},
   ISSN = {1546-170X},
   DOI = {10.1038/s41591-024-02857-3},
   year = {2024},
   type = {Journal Article}
}

@article{weinstein2013cancer,
  title     = {The cancer genome atlas pan-cancer analysis project},
  author    = {Weinstein, John N and Collisson, Eric A and Mills, Gordon B and Shaw, Kenna R and Ozenberger, Brad A and Ellrott, Kyle and Shmulevich, Ilya and Sander, Chris and Stuart, Joshua M},
  journal   = {Nature genetics},
  volume    = {45},
  number    = {10},
  pages     = {1113--1120},
  year      = {2013},
  publisher = {Nature Publishing Group}
}

@article{bulten2022artificial,
  title     = {Artificial intelligence for diagnosis and Gleason grading of prostate cancer: the PANDA challenge},
  author    = {Bulten, Wouter and Kartasalo, Kimmo and Chen, Po-Hsuan Cameron and Str{\"o}m, Peter and Pinckaers, Hans and Nagpal, Kunal and Cai, Yuannan and Steiner, David F and Van Boven, Hester and Vink, Robert and others},
  journal   = {Nature medicine},
  volume    = {28},
  number    = {1},
  pages     = {154--163},
  year      = {2022},
  publisher = {Nature Publishing Group US New York}
}

@article{ling2025camelyon,
   author = {Ling, Xitong and Lei, Yuanyuan and Li, Jiawen and Cheng, Junru and Huang, Wenting and Guan, Tian and Guan, Jian and He, Yonghong},
   title = {Comprehensive Benchmark Dataset for Pathological Lymph Node Metastasis in Breast Cancer Sections},
   journal = {Scientific Data},
   volume = {12},
   number = {1},
   pages = {1381},
   ISSN = {2052-4463},
   DOI = {10.1038/s41597-025-05586-5},
   year = {2025},
   type = {Journal Article}
}

@inproceedings{zhang2024attention,
  title={Attention-challenging multiple instance learning for whole slide image classification},
  author={Zhang, Yunlong and Li, Honglin and Sun, Yunxuan and Zheng, Sunyi and Zhu, Chenglu and Yang, Lin},
  booktitle={European conference on computer vision},
  pages={125--143},
  year={2024},
  organization={Springer}
}

@inproceedings{li2021dual,
  title={Dual-stream multiple instance learning network for whole slide image classification with self-supervised contrastive learning},
  author={Li, Bin and Li, Yin and Eliceiri, Kevin W},
  booktitle={Proceedings of the IEEE/CVF Conference on Computer Vision and Pattern Recognition},
  pages={14313--14323},
  year={2021}
}

@inproceedings{xiang2023exploring,
  title     = {Exploring Low-Rank Property in Multiple Instance Learning for Whole Slide Image Classification},
  author    = {Xiang, Jinxi and Wang, Xiyue and Zhang, Jun and Yang, Sen and Han, Xiao and Yang, Wei},
  booktitle = {The Eleventh International Conference on Learning Representations},
  year      = {2023},
  url       = {https://openreview.net/forum?id=01KmhBsEPFO}
}

@article{zimmermann2024virchow2,
  title   = {Virchow2: Scaling Self-Supervised Mixed Magnification Models in Pathology},
  author  = {Zimmermann, Eric and Vorontsov, Eugene and Viret, Julian and Casson, Adam and Zelechowski, Michal and Shaikovski, George and Tenenholtz, Neil and Hall, James and Klimstra, David and Yousfi, Razik and Fuchs, Thomas and Fusi, Nicolo and Liu, Siqi and Severson, Kristen},
  journal = {arXiv preprint arXiv:2408.00738},
  year    = {2024}
}

@article{RN7,
  author  = {Xu, Hanwen and Usuyama, Naoto and Bagga, Jaspreet and Zhang, Sheng and Rao, Rajesh and Naumann, Tristan and Wong, Cliff and Gero, Zelalem and Gonz{\'a}lez, Javier and Gu, Yu and Xu, Yanbo and Wei, Mu and Wang, Wenhui and Ma, Shuming and Wei, Furu and Yang, Jianwei and Li, Chunyuan and Gao, Jianfeng and Rosemon, Jaylen and Bower, Tucker and Lee, Soohee and Weerasinghe, Roshanthi and Wright, Bill J. and Robicsek, Ari and Piening, Brian and Bifulco, Carlo and Wang, Sheng and Poon, Hoifung},
  title   = {A whole-slide foundation model for digital pathology from real-world data},
  journal = {Nature},
  volume  = {630},
  number  = {8015},
  pages   = {181--188},
  year    = {2024}
}

@misc{filiot2024phikon_v2,
      title={Phikon-v2, A large and public feature extractor for biomarker prediction}, 
      author={Alexandre Filiot and Paul Jacob and Alice Mac Kain and Charlie Saillard},
      year={2024},
      eprint={2409.09173},
      archivePrefix={arXiv},
      primaryClass={eess.IV},
      url={https://arxiv.org/abs/2409.09173}, 
}
\bibliographystyle{iclr2027_conference}

\appendix
\section{Formal Diagnostic Derivation}
\label{app:diagnostic}

For a fixed compact-subset prediction $p$, recall
\begin{equation}
e_{\mathrm{orig}} = D(p,q_0),
\qquad
e_{\mathrm{cond}} = D(p,q_C),
\qquad
\delta_C = D(q_0,q_C).
\end{equation}
Because $D$ is a metric, the reverse triangle inequality gives
\begin{equation}
\left|e_{\mathrm{cond}}-e_{\mathrm{orig}}\right|
\leq
\delta_C .
\end{equation}
Thus, $\delta_C$ upper-bounds how much the continuous fidelity error can change solely because the prediction reference is substituted.

For the discrete FidelitySuccess criterion used in our experiments, let
\begin{equation}
\gamma_0
=
q_{0,(1)}-q_{0,(2)}
\end{equation}
denote the margin between the largest and second-largest probabilities of the original full-bag prediction. Under TV distance, if
\begin{equation}
\delta_C < \frac{\gamma_0}{2},
\end{equation}
the predicted class of $q_C$ is guaranteed to agree with that of $q_0$. In addition, the reverse-triangle bound implies that the threshold component of the fidelity verdict cannot change when
\begin{equation}
\delta_C
<
\left|e_{\mathrm{cond}}-\epsilon\right|.
\end{equation}

Defining
\begin{equation}
\rho
=
\min\left(
\frac{\gamma_0}{2},
\left|e_{\mathrm{cond}}-\epsilon\right|
\right),
\end{equation}
we therefore obtain the sufficient condition
\begin{equation}
\delta_C < \rho
\quad\Longrightarrow\quad
Z(p;q_C)=Z(p;q_0),
\end{equation}
where $Z(p;q)$ denotes the discrete FidelitySuccess verdict for subset prediction $p$ evaluated against reference $q$.

This guarantee applies only to the corresponding sample, subset prediction, and fidelity threshold. Failure to satisfy the condition means only that verdict agreement is \emph{not certified by this condition}; it does not imply disagreement or instability. The strict inequality is retained at the decision boundary. The continuous reverse-triangle bound requires the metric property of $D$, while the class-margin component above uses the TV-distance setting employed in our experiments.

\section{Datasets, Endpoints, Splits, and Cohort Provenance}
\label{app:data_scope}

\begin{table}[H]
\centering
\scriptsize
\setlength{\tabcolsep}{4pt}
\renewcommand{\arraystretch}{1.08}
\caption{\textbf{Evaluation cohorts, analysis units, and class composition.} Evaluation $N$ refers to the validation set used for the reported fidelity analyses. For PANDA and CAMELYON+, no validated estimate is included in the reported analyses, so $N$ denotes the planned validation set.}
\label{tab:dataset_scope}
\begin{tabular}{@{}p{0.21\textwidth}p{0.18\textwidth}p{0.25\textwidth}p{0.28\textwidth}@{}}
\toprule
\textbf{Task} & \textbf{Unit} & \textbf{Evaluation $N$} & \textbf{Positive class} \\
\midrule
Clean NSCLC & Patient & 95 patients (98 slides) & LUSC: 48/95 (50.5\%) \\
LUAD-TP53 & Patient & 67 of 69 validation patients & TP53 mutant: 34/67 (50.7\%) \\
CRC-MSI & Patient & 81 & MSI-H: 11/81 (13.6\%) \\
LUAD-EGFR & Patient & 46 & EGFR mutant: 6/46 (13.0\%) \\
LUAD-KRAS & Patient & 46 & KRAS mutant: 14/46 (30.4\%) \\
LUAD-STK11 & Patient & 46 & STK11 mutant: 7/46 (15.2\%) \\
BRCA morphology & Patient & 169 patients (179 slides) & IDC: 131/169 (77.5\%) \\
BRCA-ER & Patient & 103 & ER positive: 85/103 (82.5\%) \\
BRCA-PR & Patient & 102 & PR positive: 73/102 (71.6\%) \\
BRCA-HER2 & Patient & 66 & HER2 positive: 20/66 (30.3\%) \\
PANDA$^{\dagger}$ & Slide & 1,699 planned slides & Cancer: 1,257/1,699 (74.0\%) \\
CAMELYON+$^{\dagger}$ & Slide-level endpoint & 74 planned slides & Tumour: 23/74 (31.1\%) \\
\bottomrule
\end{tabular}

\vspace{2pt}
\raggedright
$^{\dagger}$No validated estimate is reported for PANDA or CAMELYON+.
\end{table}

All reported fidelity quantities use the corresponding validation partition, while held-out test partitions remain sealed and do not enter the reported analyses. The NSCLC, LUAD, CRC, and BRCA tasks are derived from TCGA cohorts~\citep{weinstein2013cancer}. Clean NSCLC uses a patient-disjoint split with seed 20260916 and 70/10/20 target fractions. LUAD-TP53 and CRC-MSI use patient-level splits with seed 20260913 and 70/15/15 fractions; the LUAD-TP53 audit includes 67 of the 69 validation patients after the documented geometry exclusion. LUAD-EGFR, LUAD-KRAS, and LUAD-STK11 use patient-level 70/10/20 splits with seed 20260916. The BRCA receptor endpoints share a patient-level 70/10/20 split generated with seed 20260913 before endpoint labels were joined.

The BRCA morphology split is patient-disjoint, but the original split seed, target ratio, and builder metadata were not recovered from the available records. PANDA~\citep{bulten2022artificial} uses a fixed slide-disjoint split because no patient identifier is available; the original split seed and the derivation of the stored binary benign/cancer label were not recovered. PANDA does not contribute a validated estimate to the breadth results reported in the main text.

For CAMELYON+~\citep{ling2025camelyon}, we use a provenance-verified subset of 875 slides rather than claiming coverage of the full benchmark. The CAMELYON16 test set is retained as the sealed test partition, while CAMELYON16-train and CAMELYON17-train form the development pool. CAMELYON17 provides patient identifiers, but an authoritative patient mapping for CAMELYON16 was not identified in the available provenance sources; we therefore do not claim patient-level disjointness for the CAMELYON16 portion. No CAMELYON+ estimate enters the reported breadth results.

\section{MIL Architectures and Pathology Encoders}
\label{app:model_scope}

The common task-breadth panel uses nine MIL architectures: MeanMIL, MaxMIL, ABMIL~\citep{ilse2018attention}, CLAM~\citep{lu2021data}, DSMIL~\citep{li2021dual}, TransMIL~\citep{shao2021transmil}, Transformer, ILRA~\citep{xiang2023exploring}, and ACMIL~\citep{zhang2024attention}.

Cross-encoder evaluation is organized around three anchor tasks---clean NSCLC, LUAD-TP53, and CRC-MSI. The planned encoder inventory comprises UNI2-h~\citep{chen2024towards}, Phikon-v2~\citep{filiot2024phikon_v2}, Virchow2~\citep{zimmermann2024virchow2}, and Prov-GigaPath~\citep{RN7}. The encoder analysis under the shared $C_{\mathrm{L2}}$ protocol uses  MeanMIL, ABMIL, TransMIL, ACMIL, and ILRA. The main-text matched panel is restricted to the eight task--MIL settings with validated estimates under both UNI2-h and Phikon-v2. Architectures requiring a different candidate-construction path are not pooled into this comparison.

Encoder comparisons require matched physical tile coordinates and footprints. Configurations that cannot satisfy this parity are excluded from matched encoder comparisons rather than harmonized through post hoc re-tiling.

\section{Candidate Construction and Ranking Specifications}
\label{app:candidate_specs}

Random candidate construction and Random-$K$ subset selection use deterministic identity-keyed seeds so that sampled indices are invariant to row ordering and cohort filtering under an otherwise fixed experimental condition. The implemented effective budgets are $M_c=\min(M,N)$ and $K_c=\min(K,M_c)$. When $N<M$, both $C_{\mathrm{L2}}$ and $C_{\mathrm{RAND}}$ coincide with the full bag; no padding, replacement sampling, or sample exclusion is applied.

Candidate scores for $C_{\mathrm{L2}}$ are computed as $\lVert x_i\rVert_2$, where $x_i$ is the predictor-specific token embedding for tile $i$ before attention, pooling, or transformer contextualization. ABMIL, TransMIL, Transformer, ILRA, CLAM, and DSMIL include a two-dimensional positional term derived from slide-wise min--max-normalized coordinates, whereas MeanMIL, MaxMIL, and ACMIL do not use coordinates in this embedding. Thus, the score does not depend on other tiles' feature values, although coordinate-using models depend on the slide-level coordinate normalization range.

Predictor-native ranking is evaluated for ABMIL, ACMIL, CLAM, and TransMIL. For each slide, the native score for every tile is computed once from the full bag using the same frozen predictor that produces the reference and subset predictions; within each candidate pool, the $K_c$ tiles with the highest scores are then selected. For ABMIL and CLAM, the native score is the pre-softmax gated-attention logit from the single attention branch. For ACMIL, it is the arithmetic mean of the pre-softmax gated-attention logits across its five attention branches. For TransMIL, it is the dot product between each pre-encoder tile-token embedding and the learned class-token vector. These scores are class-agnostic, are not normalized, and do not depend on $K$ except through the number of selected tiles. Exact ties are not assigned an additional explicit tie-breaking rule.

\section{Fidelity Evaluation and Statistical Analysis}
\label{app:stats}

Statistical inference follows a patient-first resampling strategy whenever patient identifiers are available. Predictor seeds are kept fixed as repeated model realizations and are not counted as independent patient-level observations. For PANDA, which lacks patient identifiers, the evaluation protocol uses slide-level resampling.

For pooled five-seed estimates, all binary fidelity outcomes from every slide--seed record belonging to a patient are averaged with equal record weight to form one patient-level rate for each candidate and reference. Cohort-level success rates then average these patient-level rates equally over the common patient set evaluated for both $C_{\mathrm{L2}}$ and $C_{\mathrm{RAND}}$. Pairwise $\Delta_{\mathrm{orig}}$, $\Delta_{\mathrm{cond}}$, and $R$ are recomputed jointly in 2,000 paired bootstrap resamples of patients with replacement, and pointwise percentile 95\% confidence intervals are reported; these are not simultaneous family-wise intervals over the full grid.

For comparisons between candidate construction strategies, the direct pairwise reference effect $R_{A,B}$ is estimated by bootstrap rather than inferred from separate significance tests under the two references. Model selection and experimental decisions are made without using the held-out test sets.

As a sensitivity check on the fixed magnitude tolerance, the 15-comparison Random-$K$ panel yielded \textsc{Reversed}/\textsc{Preserved}/\textsc{Ranking-Indeterminate} counts of 6/3/6, 6/3/6, 5/3/7, 4/2/9, 3/2/10, and 3/0/12 for $\tau=0$, $0.04$, $0.045$, $0.05$, $0.07$, and $0.10$, respectively. Under a stricter variant requiring the entire sign-consistent 95\% confidence interval to lie beyond $|\Delta|=0.05$, the counts were 2/0/13. The main analysis retains the fixed $\tau=0.05$ rule.

Because $B_g$ is an aggregate change in success rate, a value near zero does not imply that individual sample-level fidelity verdicts are unchanged: opposing sample-level changes may cancel in aggregate.

\section{Provenance and Validation Controls}
\label{app:provenance}

All reported results are linked to their feature source, data split, predictor checkpoint, and evaluation configuration. Results that fail split, seed, feature-extraction, or provenance checks are excluded from the reported analyses. Protocol-incompatible configurations are excluded from matched comparisons rather than harmonized through post hoc preprocessing or cohort redefinition.

\section{Additional Numerical Results and Breadth Inventory}
\label{app:additional_results}
\FloatBarrier

\begin{figure}[!ht]
\centering
\includegraphics[width=\linewidth]{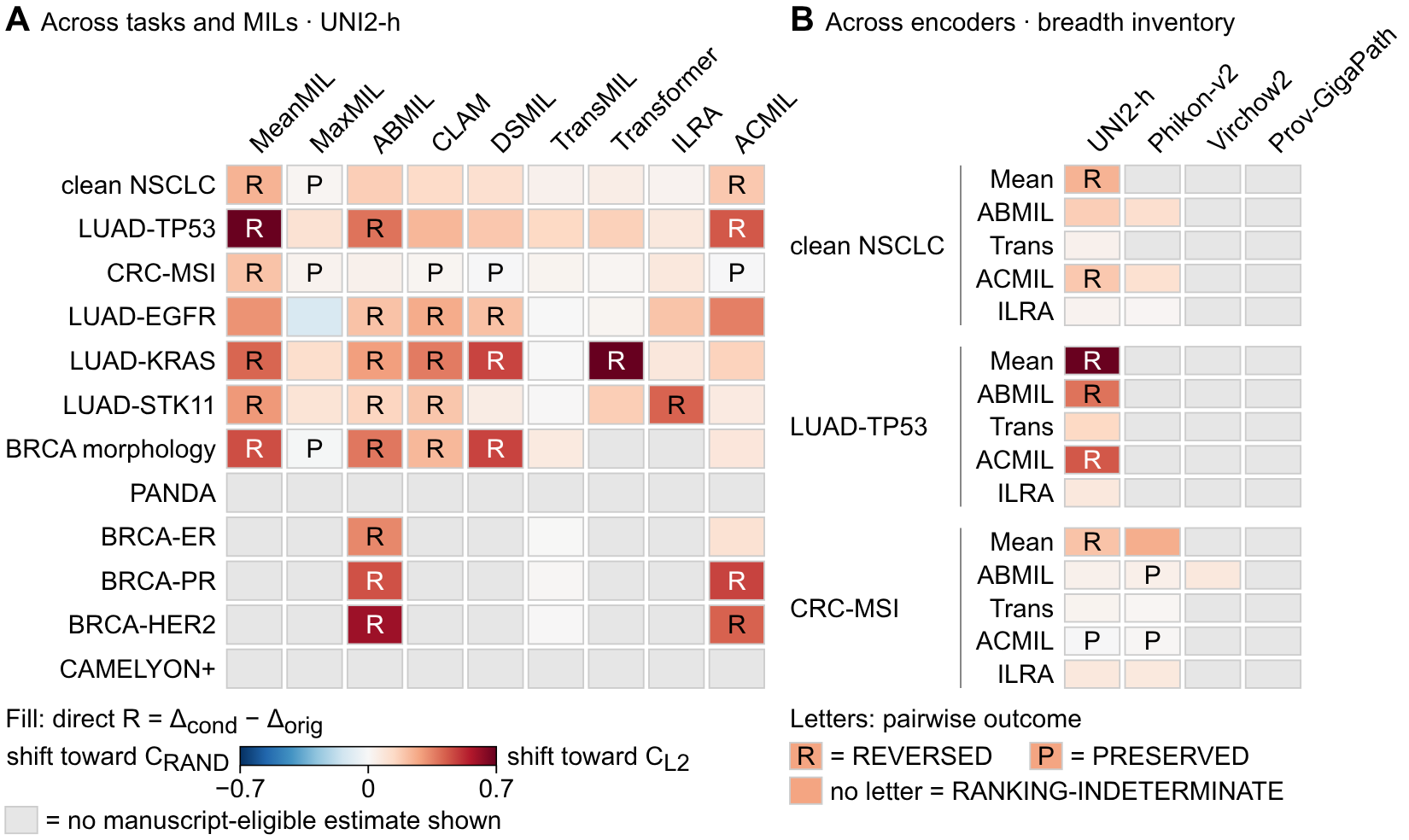}
\caption{\textbf{Full breadth-audit inventory underlying the scoped main-text Figure~3.}
Panel A shows the full task--MIL inventory for the common-$C_{\mathrm{L2}}$ audit, and Panel B shows the full encoder-breadth inventory over the planned anchor task--MIL settings. Colored cells denote validated estimates included in the reported analyses, with fill indicating direct $R=\Delta_{\mathrm{cond}}-\Delta_{\mathrm{orig}}$ and letters indicating the categorical pairwise outcome. Grey cells denote configurations without a validated estimate. The main-text Figure~3 presents complete common-coverage and encoder-matched subsets for compact comparison.}
\label{fig:app_full_inventory}
\end{figure}

In the additional CRC-MSI/ABMIL three-encoder anchor, Virchow2 yielded $R=0.0765$ [0.0321, 0.1235] and remained \textsc{Ranking-Indeterminate}; the corresponding UNI2-h and Phikon-v2 estimates are shown in Fig.~\ref{fig:app_full_inventory}.

Table~\ref{tab:app-p3p4-full} reports the complete five-seed pooled comparisons underlying the 15-comparison Random-$K$ audit and the matched native-ranking stress test in the main text.

\begin{sidewaystable}
\centering
\small
\setlength{\tabcolsep}{3pt}
\caption{\textbf{Full numerical results for the main Random-$K$ audit and matched native-ranking stress test.} Results compare $C_{\mathrm{L2}}$ against $C_{\mathrm{RAND}}$ at the default operating point ($M=1024$, $K=32$, $\epsilon=0.05$) using UNI2-h features. Each row pools five predictor seeds ($0,1,42,1337,2025$) with patient-first bootstrap inference. Intervals are percentile 95\% confidence intervals from 2,000 bootstrap replicates. The Random-$K$ rows comprise the 15 five-seed comparisons reported in the main audit; the native-ranking rows comprise the four matched clean-NSCLC stress-test comparisons. Status denotes whether the direct $R$ interval excludes zero, and Outcome denotes the categorical pairwise result under the fixed tolerance rule ($\tau=0.05$).}
\label{tab:app-p3p4-full}
\begin{tabular}{@{}lllrcccll@{}}
\toprule
Task & MIL & Ranker & $n$ & $\Delta_{\mathrm{orig}}$ [95\% CI] & $\Delta_{\mathrm{cond}}$ [95\% CI] & direct $R$ [95\% CI] & Status & Outcome \\
\midrule
CRC-MSI & ABMIL & Random-K & 81 & $0.0494$ [$0.0049$, $0.0938$] & $0.084$ [$0.042$, $0.1259$] & $0.0346$ [$-0.0$, $0.0741$] & ns & \texttt{RANKING\_INDETERMINATE} \\
CRC-MSI & ACMIL & Random-K & 81 & $0.0914$ [$0.0519$, $0.1333$] & $0.0864$ [$0.0469$, $0.1309$] & $-0.0049$ [$-0.0297$, $0.0222$] & ns & \texttt{PRESERVED} \\
CRC-MSI & TransMIL & Random-K & 81 & $0.0074$ [$-0.0173$, $0.0321$] & $0.0296$ [$0.0$, $0.0617$] & $0.0222$ [$0.0$, $0.0519$] & ns & \texttt{RANKING\_INDETERMINATE} \\
LUAD-TP53 & ABMIL & Random-K & 67 & $-0.1881$ [$-0.2597$, $-0.1134$] & $0.194$ [$0.1164$, $0.2746$] & $0.3821$ [$0.2955$, $0.4627$] & SIG & \texttt{REVERSED} \\
LUAD-TP53 & ACMIL & Random-K & 67 & $-0.3284$ [$-0.391$, $-0.2597$] & $0.1075$ [$0.0269$, $0.1851$] & $0.4358$ [$0.3552$, $0.5104$] & SIG & \texttt{REVERSED} \\
LUAD-TP53 & TransMIL & Random-K & 67 & $-0.0985$ [$-0.1434$, $-0.0507$] & $0.0448$ [$-0.0209$, $0.1104$] & $0.1433$ [$0.0806$, $0.209$] & SIG & \texttt{RANKING\_INDETERMINATE} \\
Clean NSCLC & ABMIL & Random-K & 95 & $-0.1232$ [$-0.1863$, $-0.0684$] & $0.0495$ [$0.0032$, $0.0968$] & $0.1726$ [$0.1116$, $0.2463$] & SIG & \texttt{RANKING\_INDETERMINATE} \\
Clean NSCLC & ACMIL & Random-K & 95 & $-0.1232$ [$-0.1842$, $-0.0737$] & $0.0663$ [$0.0221$, $0.1158$] & $0.1895$ [$0.1242$, $0.2674$] & SIG & \texttt{REVERSED} \\
Clean NSCLC & CLAM & Random-K & 95 & $-0.0516$ [$-0.1137$, $0.0032$] & $0.0832$ [$0.0379$, $0.1306$] & $0.1347$ [$0.0779$, $0.2042$] & SIG & \texttt{RANKING\_INDETERMINATE} \\
Clean NSCLC & DSMIL & Random-K & 95 & $-0.0442$ [$-0.0822$, $-0.0063$] & $0.0695$ [$0.0295$, $0.1116$] & $0.1137$ [$0.0737$, $0.16$] & SIG & \texttt{RANKING\_INDETERMINATE} \\
Clean NSCLC & ILRA & Random-K & 95 & $0.0105$ [$-0.0064$, $0.0295$] & $0.0337$ [$0.0147$, $0.0547$] & $0.0232$ [$0.0021$, $0.0463$] & SIG & \texttt{RANKING\_INDETERMINATE} \\
Clean NSCLC & MaxMIL & Random-K & 95 & $0.0821$ [$0.0442$, $0.1221$] & $0.0968$ [$0.0589$, $0.1368$] & $0.0147$ [$-0.0105$, $0.0421$] & ns & \texttt{PRESERVED} \\
Clean NSCLC & MeanMIL & Random-K & 95 & $-0.1379$ [$-0.1916$, $-0.0905$] & $0.1049$ [$0.0579$, $0.1509$] & $0.2428$ [$0.1747$, $0.3095$] & SIG & \texttt{REVERSED} \\
Clean NSCLC & TransMIL & Random-K & 95 & $-0.0411$ [$-0.0789$, $-0.0063$] & $-0.0084$ [$-0.0463$, $0.0274$] & $0.0326$ [$0.0147$, $0.0526$] & SIG & \texttt{RANKING\_INDETERMINATE} \\
Clean NSCLC & Transformer & Random-K & 95 & $-0.0705$ [$-0.1147$, $-0.0295$] & $-0.0221$ [$-0.0705$, $0.0242$] & $0.0484$ [$0.0189$, $0.0821$] & SIG & \texttt{RANKING\_INDETERMINATE} \\
\midrule
Clean NSCLC & ABMIL & native & 95 & $0.0337$ [$-0.0042$, $0.0716$] & $0.1821$ [$0.1211$, $0.2474$] & $0.1484$ [$0.0884$, $0.2179$] & SIG & \texttt{RANKING\_INDETERMINATE} \\
Clean NSCLC & ACMIL & native & 95 & $0.0744$ [$0.0358$, $0.1144$] & $0.2428$ [$0.1754$, $0.3144$] & $0.1684$ [$0.1073$, $0.2421$] & SIG & \texttt{PRESERVED} \\
Clean NSCLC & CLAM & native & 95 & $-0.0105$ [$-0.0358$, $0.0105$] & $0.1126$ [$0.0611$, $0.1705$] & $0.1232$ [$0.0674$, $0.1874$] & SIG & \texttt{RANKING\_INDETERMINATE} \\
Clean NSCLC & TransMIL & native & 95 & $0.0084$ [$-0.0274$, $0.0484$] & $0.0263$ [$-0.0116$, $0.0663$] & $0.0179$ [$0.0$, $0.0379$] & ns & \texttt{RANKING\_INDETERMINATE} \\
\bottomrule
\end{tabular}
\end{sidewaystable}

\end{document}